\documentclass[conference]{IEEEtran}
\IEEEoverridecommandlockouts
\usepackage{cite}
\usepackage{amsmath,amssymb,amsfonts}
\usepackage{graphicx}
\usepackage{textcomp}
\usepackage{xcolor}
\usepackage{booktabs}
\usepackage{array}
\usepackage{makecell}
\usepackage{url}
\def\BibTeX{{\rm B\kern-.05em{\sc i\kern-.025em b}\kern-.08em
    T\kern-.1667em\lower.7ex\hbox{E}\kern-.125emX}}
\begin{document}

\title{CyberAId: AI-Driven Cybersecurity for Financial Service Providers
\\
\thanks{This work was supported by the European Union under the DIGITAL EUROPE Programme, grant agreement No.~101249596 (CyberAId).}
}

\author{
\IEEEauthorblockN{
George Fatouros\IEEEauthorrefmark{1},
Georgios Makridis\IEEEauthorrefmark{2},
John Soldatos\IEEEauthorrefmark{1},
Dimosthenis Kyriazis\IEEEauthorrefmark{2},
Pedro Malo\IEEEauthorrefmark{5},\\
George Kousiouris\IEEEauthorrefmark{3},
Giannis Ledakis\IEEEauthorrefmark{4},
Louiza Kachrimani\IEEEauthorrefmark{4},
Panagiotis Rizomiliotis\IEEEauthorrefmark{3}
Bruno Almeida\IEEEauthorrefmark{5},\\
Despina Tomkou\IEEEauthorrefmark{1},
Kostas Metaxas\IEEEauthorrefmark{6},
Konstantinos Ilias\IEEEauthorrefmark{6},
Christos Gkizelis\IEEEauthorrefmark{7},
Ernstjan de Gooyert\IEEEauthorrefmark{8},
Amin Babazadeh\IEEEauthorrefmark{8},\\
Kostis Mavrogiorgos\IEEEauthorrefmark{2},
Pepi Paraskevoulakou\IEEEauthorrefmark{2},
Christos Xenakis\IEEEauthorrefmark{9},
Giannis Chouchoulis\IEEEauthorrefmark{9},
Konstantina Tripodi\IEEEauthorrefmark{10}}
\IEEEauthorblockA{
\IEEEauthorrefmark{1}\textit{Innov-Acts}, Cyprus; \quad
\IEEEauthorrefmark{2}\textit{University of Piraeus}, Greece; \\
\IEEEauthorrefmark{3}\textit{Harokopio University of Athens}, Greece; \quad
\IEEEauthorrefmark{4}\textit{Ubitech}, Cyprus;\quad
\IEEEauthorrefmark{5}\textit{Unparallel Innovation}, Portugal; \\
\IEEEauthorrefmark{6}\textit{KM Cube Investment Services}, Greece; \quad
\IEEEauthorrefmark{7}\textit{Hellenic Telecommunications Organisation}, Greece; \\ 
\IEEEauthorrefmark{8}\textit{Qubo Technology}, Austria; \quad
\IEEEauthorrefmark{9}\textit{InQbit}, Romania; \quad
\IEEEauthorrefmark{10}\textit{JRC Capital Management}, Germany \\[0.5em]
\{gfatouros, jsoldatos, dtomkou\}@innov-acts.com;\;
\{gmakridis, dimos, komav, e.paraskevoulakou\}@unipi.gr;\; \\
\{gkousiou, prizomil\}@hua.gr;
\{gledakis, lkachrimani\}@ubitech.eu;\;
\{pedro.malo, bruno.almeida\}@unparallel.pt;\;\\
\{kostas, k.ilias\}@km3.gr;\;
cgkizelis@ote.gr;
\{ernstjan.degooyert,amin.babazadeh\}@qubo.technology;\;\\
\{chris, giannis.chouchoulis\}@inqbit.io;\;
ktripodi@jrconline.com;
}
}

\maketitle

\begin{abstract}
European financial institutions face mounting regulatory pressure under DORA, NIS2 and the EU AI Act while their security operations centres remain constrained not by data or staffing but by reasoning capacity: enterprise SIEMs cover only a fraction of MITRE ATT\&CK techniques, two thirds of SOC teams cannot keep pace with alert volumes, and the majority of breaches are preceded by alerts that are generated but never investigated. Frontier large language models now achieve state-of-the-art results on isolated cybersecurity tasks — one-day vulnerability exploitation, code-level patching, intrusion detection — yet no narrow win constitutes a platform that can compose across functions, persist multi-tenant state, map findings to regulatory regimes and survive an audit. This position paper argues that the right unit of construction is a hybrid multi-agent system in which specialised LLM subagents reason over classical SIEM/XDR telemetry rather than replacing it, share accumulated agent state across institutions through privacy-preserving federation, and can connect to complementary capability packs such as quantum-based authentication, digital twins for adversarial validation, and eBPF-based kernel telemetry. We present CyberAId, a model-agnostic, on-premise-deployable platform in which a Main Agent/CRA coordination layer, a Reporting capability, and specialist subagents operate within a shared runtime under bounded human-in-the-loop autonomy, organised around four falsifiable design principles, and aligned with DORA reporting timelines and the EU AI Act. CyberAId will be validated at four representative financial use cases — client impersonation, anti-money-laundering for payment service providers, retail-banking incident response, and high-frequency-trading resilience — and propose skill-based agent adaptation as the most promising research direction for turning each deployment into a contribution to a continuously refined collective defence.
\end{abstract}

\begin{IEEEkeywords}
LLM agents, multi-agent systems, financial cybersecurity, retrieval-augmented generation, DORA, digital twins, cyberaid.
\end{IEEEkeywords}

\section{Introduction}\label{sec:intro}

The European financial sector manages over €100~trillion in assets and processes more than 400{,}000 transactions per day, while absorbing a 238\% growth in cyberattacks since 2020 and roughly three times the attack volume of the next-most-targeted industry~\cite{ref-eba-ai,ref-ibm-breach}.
The threats are concrete and increasingly automated: \emph{client impersonation} via compromised email and endpoint devices, \emph{payment and transaction fraud} that stays below detection thresholds, \emph{money laundering} via structuring and layering across e-money and card-payment ecosystems, \emph{account takeover} and credential stuffing, \emph{market manipulation} on trading platforms, \emph{insider threats} that signature-based controls cannot surface, and \emph{supply-chain compromise} through the thousands of third-party connections that characterise modern finance.
Each of these scenarios spans data sources, business processes, and regulatory regimes that no single security tool was designed to cover.

\subsection{The reasoning bottleneck}\label{sec:intro-bottleneck}

Financial security operations centres (SOCs) are increasingly limited not by data or staffing, but by reasoning capacity.
Industry surveys report that enterprise SIEMs detect only $\sim$21\% of MITRE ATT\&CK techniques despite ingesting data that could in principle cover the great majority, that around two thirds of SOC teams cannot keep pace with alert volumes, and that the majority of breaches are preceded by alerts that were generated but never investigated~\cite{ref-soc-survey}.
The consequence is structural~\cite{ref-ibm-breach}: dwell times for advanced persistent threats remain measured in months unless reasoning across heterogeneous signals is automated.

\subsection{Why narrow LLM wins do not generalise}\label{sec:intro-narrow}

LLMs have begun to deliver state-of-the-art on isolated cybersecurity tasks: frontier models autonomously exploit 87\% of one-day Common Vulnerabilities and Exposures (CVEs) from a textual description~\cite{ref-llm-exploits}; multi-agent teams achieve a $4.3\times$ improvement over single-agent baselines on zero-day exploitation~\cite{zhu2026teams}; LLM-assisted static analysis raises vulnerability-detection F1 above traditional SAST tools~\cite{ref-iris}; and domain-specific systems reach 0.97 F1 on intrusion detection~\cite{ref-idsagent}.

\emph{None of these is a platform.}
A SOC platform must compose across functions, persist multi-tenant state, ingest live threat intelligence, map findings to the regulatory regimes that govern financial services, generate audit-grade artefacts, and survive adversarial probing of the agents themselves.
A monolithic LLM bolted onto an existing SIEM does none of these things.
The right unit is a \emph{hybrid system}: classical SIEM/XDR for ground truth and high-throughput telemetry, specialised LLM agents for reasoning, and architectural mechanisms for trust.

\subsection{The CyberAId vision}\label{sec:intro-vision}

CyberAId is conceived as an \emph{integrated, hybrid AI-cyber\-security platform for financial critical infrastructures} that coordinates state-of-the-art security tools through a shared runtime of logical LLM subagents, federates accumulated knowledge across institutions, can integrate complementary capability packs (quantum token authentication, digital twins for adversarial validation, eBPF-based kernel telemetry), and operates under bounded human-in-the-loop autonomy.
The intended outcome is a system installed inside the institution, not a cloud copilot.

\subsection{Contributions}\label{sec:intro-contributions}

The paper makes four contributions: a \emph{position} on hybrid LLM-agent / SIEM-XDR cybersecurity for finance (Section~\ref{sec:background}--\ref{sec:position}); an \emph{agent-runtime architecture} with one Main Agent/CRA coordination layer, a Reporting capability, and specialist subagents covering the financial cyber-resilience lifecycle (Section~\ref{sec:platform}); a \emph{methods stack} combining partitioned RAG, federated cross-institutional knowledge sharing, and optional capability packs for digital-twin validation, eBPF telemetry and quantum token authentication (Section~\ref{sec:methods}); and four \emph{financial use cases} that exercise the platform end-to-end (Section~\ref{sec:usecases}), followed by a discussion of \emph{skill-based agent adaptation} as a research direction (Section~\ref{sec:discussion}).

\section{Background and Related Work}\label{sec:background}

The literature directly relevant to CyberAId clusters around three observations: LLM agents are now usable building blocks; hybrid pipelines outperform end-to-end LLM pipelines on the metrics that matter to a SOC; and the financial-sector application is almost untouched despite imminent regulatory pressure.

\subsection{LLM agents as building blocks}\label{sec:bg-agents}

Two design patterns now dominate practical agent construction: \emph{ReAct}-style tool-calling that interleaves reasoning with tool invocation~\cite{ref-react}, and \emph{plan-and-execute} that decomposes tasks before acting~\cite{ref-plan-exec}.
Both compose into multi-agent orchestrations through frameworks such as AutoGen~\cite{ref-autogen}, with retrieval-augmented generation, chain-of-thought prompting~\cite{wei2022chain, fatouros2026signal} and programmatic tool use~\cite{schick2023toolformer} providing the substrate.
A consistent pattern across cybersecurity-specific systems is that \emph{orchestration quality matters more than model scale}: a small, well-orchestrated multi-agent system can outperform a single large model on incident-response actionability by a wide margin~\cite{drammeh2025multi}, and hierarchical multi-agent attack teams achieve a $4.3\times$ improvement over single-agent baselines on zero-day exploitation~\cite{zhu2026teams}.

\subsection{Hybrid LLM pipelines outperform end-to-end ones}\label{sec:bg-hybrid}

The second observation is that grounding LLM reasoning in classical detection, retrieval or planning consistently improves accuracy, latency and auditability over end-to-end neural pipelines.
SIEM-grounded triage agents~\cite{ref-cortex}, agentic-RAG attack-classification systems~\cite{blefari2025cyberrag}, and explainable intrusion-detection agents~\cite{ref-idsagent} all retain a deterministic detection floor and use the LLM for cross-domain synthesis, alert enrichment and natural-language interaction.
On the threat-intelligence side, graph-augmented agents over CVE, CWE and ATT\&CK chains~\cite{ref-ctikg} produce structured CTI artefacts at consistencies that are difficult to reach through manual curation; phishing detection follows the same arc, with knowledge-graph- and debate-augmented detectors outperforming single-pass LLM classifiers~\cite{ref-knowphish}.
The throughline is consistent: an LLM reasoning over a known-good telemetry surface, with a known-good knowledge partition, is more accurate and more auditable than the same model operating end-to-end.

\subsection{Two gaps that motivate CyberAId}\label{sec:bg-gaps}

Two gaps in this literature directly motivate the position taken here.

\paragraph{Financial cybersecurity is almost untouched}
Despite the regulatory pressure of DORA, NIS2 and the EU AI Act~\cite{ref-dora,ref-nis2,ref-aiact} and EBA reporting that a majority of EU payment institutions are adopting AI~\cite{ref-eba-ai}, peer-reviewed work on LLM agents \emph{specifically targeting financial cybersecurity} is sparse: the closest dedicated study addresses STRIDE-based banking threat-modelling~\cite{wu2024threatmodeling}, and to the best of our knowledge no work proposes LLM-agent automation for DORA or NIS2 compliance workflows, financial SOC operations under regulatory reporting timelines, or cross-institutional federated defence.
CyberAId therefore builds on the agent literature in cybersecurity rather than on a body of work in finance.

\paragraph{Agent-level adversarial robustness is rarely first-class}
Recent surveys catalogue an expanding threat surface against agent systems: prompt injection, knowledge-base poisoning, tool-call hijacking, and protocol-level exploits~\cite{ref-survey-llm-sec,ferrag2025prompt}.
Knowledge-base poisoning attacks reach high success rates against unprotected RAG, and dedicated agent-targeted defences remain a minority concern~\cite{ref-redteamllm}.
For a regulated financial deployment, agent-level adversarial robustness must be an architectural requirement, not an afterthought; this directly motivates principle P4 in Section~\ref{sec:position}.

\section{Position and Design Principles}\label{sec:position}

CyberAId is built around four principles, restated here as falsifiable claims for future work to refute or refine.

\noindent\textbf{P1.~Hybrid grounding.}
Agents \emph{reason over} SIEM/XDR telemetry rather than replacing it.
The LLM layer provides cross-domain synthesis, regulatory mapping, and natural-language interaction; classical tooling provides high-throughput ingestion, deterministic detection on signatures, and a stable ground truth that the agent can be held against.
Outputs flow back into the SIEM as detection rules, enrichment fields, and Security Orchestration, Automation, and Response (SOAR)  playbooks~\cite{blefari2025cyberrag,ref-cortex}.

\noindent\textbf{P2.~Specialisation and composition.}
Domain expertise is partitioned across specialist subagents, each with its own system instructions, knowledge partition, tool set, and Finding Schema.
Composition is explicit and auditable: one Main Agent/CRA coordination layer aggregates findings, resolves conflicts and synthesises recommendations, while a Reporting capability provides a direct query path to specialists for compliance and incident reporting.

\noindent\textbf{P3.~Federated knowledge.}
The unit shared across institutions is \emph{accumulated agent knowledge}~--~playbooks, skills, behavioural baselines, critical findings from agent runtime~--~not raw data.
Federated learning, secure multi-party computation, and differential privacy let smaller institutions inherit detection capability from larger participants without exporting customer data, replacing manual sectoral threat-intelligence sharing with continuous machine-mediated exchange.

\noindent\textbf{P4.~Architectural trust.}
Trust is engineered, not assumed.
Bounded autonomy with tiered human-in-the-loop (HITL) gates, partitioned RAG scoped by a Security Context object, direct/CLI-first tool integration with deterministic schemas, append-only audit, prompt-injection and output-sanitisation guardrails, and dynamic skill-based adaptation together form a trust envelope aligned with DORA's reporting timelines and Article~14 of the EU AI Act.

\section{The CyberAId Multi-Agent Platform}\label{sec:platform}

\begin{figure*}[!t]
\centering
\includegraphics[width=0.92\textwidth]{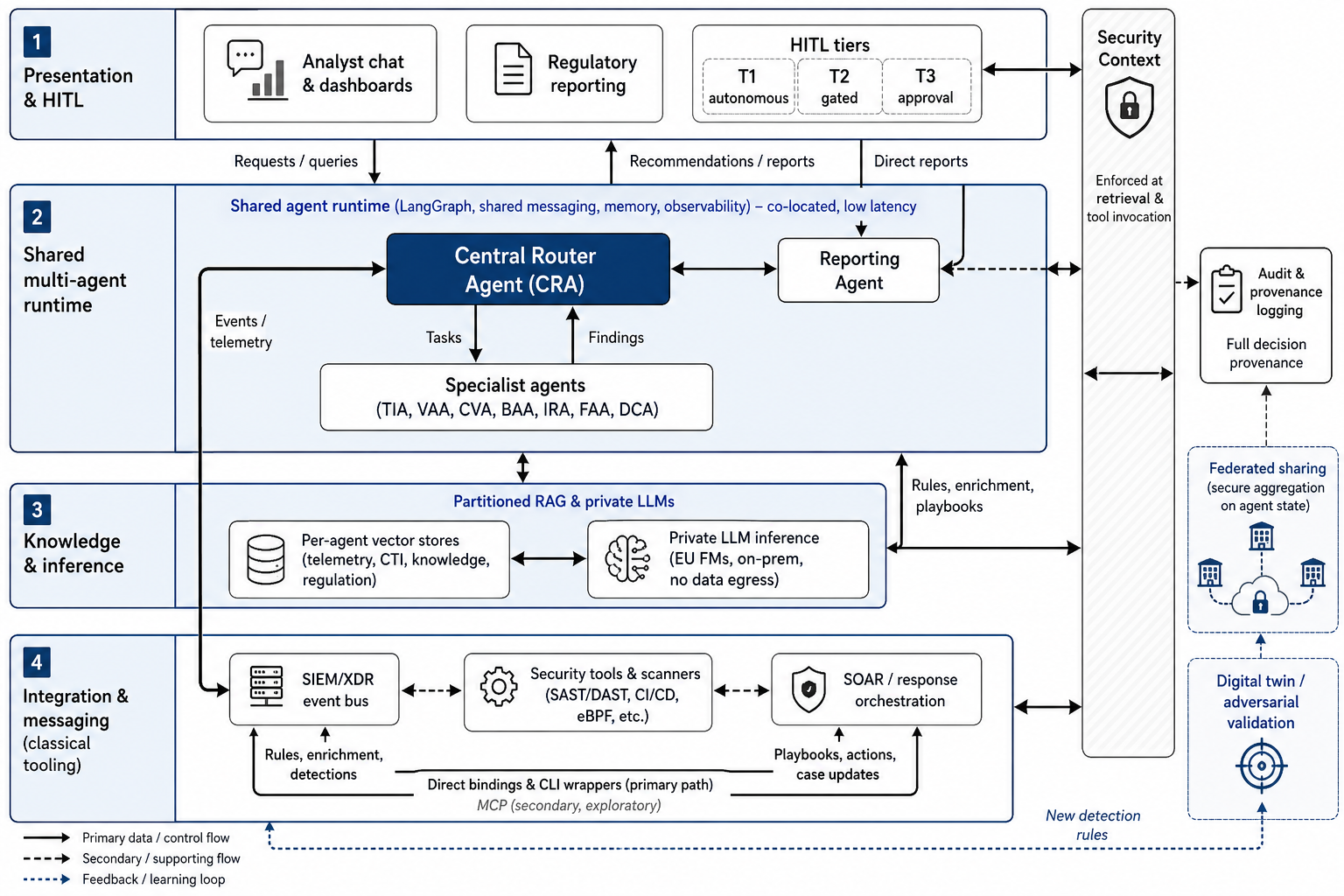}
\caption{Conceptual architecture of the CyberAId platform. Specialist LLM agents share a single multi-agent runtime that reasons over classical SIEM/XDR telemetry; a Security Context object enforced at retrieval and tool-invocation time provides the trust envelope; outputs are written back to the SIEM as detection rules and SOAR playbooks/skills.}
\label{fig:architecture}
\end{figure*}

This section focuses on the conceptual architecture of the platform.
As shown in Fig.~\ref{fig:architecture}, the platform is layered: Integration and Messaging at the edge, a Domain Agents layer in which all specialists co-exist within a single agent runtime (e.g.\ LangGraph\footnote{https://github.com/langchain-ai/langgraph}) with shared messaging, memory and observability, a Knowledge and Inference layer with partitioned vector stores and the LLM inference service, and a Presentation/HITL layer for analyst chat and executive dashboards.
Co-locating the agents in one runtime keeps inter-agent latency low and gives the Main Agent/CRA coordination layer direct access to specialist state for conflict resolution.
Tool integration is therefore primarily through direct library calls and CLI wrappers (Section~\ref{sec:methods-tools}); the Model Context Protocol (MCP) is retained as an optional external integration protocol behind the Tool Adapter Layer, not as the agent transport.
A Security Context object created at the API gateway flows through every agent boundary and is enforced at retrieval and tool-invocation time.

\subsection{Specialist agent roster}\label{sec:agents}

A Main Agent/CRA coordination layer, a Reporting capability, and specialist subagents run within a shared agent runtime, with each specialist retaining its own system instructions, knowledge partition, tool bindings, and standardised Finding Schema.
The composition mirrors the division of labour observed in mature security teams.

\paragraph{Main Agent / Central Router Agent (CRA)}
Acts as the analyst-facing triage and dispatch layer.
Receives normalised events from the SIEM event bus, performs initial severity assessment and impact scoring, activates the appropriate specialists in parallel, aggregates structured findings, resolves conflicts using calibrated confidence scores, and synthesises a final recommendation with full chain-of-thought provenance.
The CRA is also the integration point for HITL escalation and the entry point for autonomous CRA orchestration patterns (sequential, peer, debate, scheduled) drawn from the multi-agent literature~\cite{ref-autogen}.

\paragraph{Reporting capability}
Provides a direct, lower-latency query path to any specialist agent without routing through the CRA, so compliance reporting and incident-response operations can run concurrently without contention.
Templates regulatory artefacts (DORA 4-hour notifications, NIS2 24-hour early warnings, MiFID~II incident reports) and assembles audit packages with traceable evidence references.

\paragraph{Threat Intelligence Agent (TIA)}
Tracks indicators of compromise and TTPs (Tactics, Techniques, and Procedures), maps observations to MITRE ATT\&CK~\cite{ref-attack}, attributes activity to known actors, and correlates campaigns across CTI feeds (STIX/TAXII, vendor advisories).
Knowledge is grounded against a graph-augmented retrieval store spanning CVE/NVD, CWE, CAPEC and ATT\&CK.

\paragraph{Vulnerability Assessment Agent (VAA)}
Scores exploitability and impact against the institution's actual exposure profile, cross-references patch availability, and prioritises remediation under business-impact and regulatory constraints.
Integrates SAST, DAST and LLM-assisted static analysis, and can consume results from optional adversarial validation in digital twins (Section~\ref{sec:methods-pentest}).

\paragraph{Compliance Verification Agent (CVA)}
Maintains live awareness of DORA, NIS2, GDPR, PSD2, 4AML and the EU AI Act~\cite{ref-dora,ref-nis2,ref-aiact}, alongside NIST CSF, ISO/IEC 27001 and PCI-DSS~v4.0.
Maps Findings to specific control objectives, generates audit-ready evidence packages, and produces continuously updated regulatory dashboards with control-by-control status.

\paragraph{Behavioural Analysis Agent (BAA)}
Runs UEBA logic against established baselines, surfaces insider threats, and flags behavioural deviations that signature-based detection cannot reach.
Operates over kernel-level telemetry (Section~\ref{sec:methods-ebpf}), authentication patterns and transaction flows, with confidence-scored output back to the CRA.

\paragraph{Incident Response Agent (IRA)}
Generates structured containment playbooks, coordinates cross-system isolation actions, manages recovery sequencing with full state tracking, and drafts regulatory notifications.
Tightly coupled with the Forensic Analysis Agent for evidence preservation during containment.

\paragraph{Forensic Analysis Agent (FAA)}
Reconstructs event timelines from heterogeneous logs (system, application, network, transaction), preserves evidence chains, and produces findings in formats that meet legal and regulatory standards for digital forensics.
Output is admissible-grade by construction.

\paragraph{DevSecOps and Code Analysis Agent (DCA)}
Performs static and dynamic code analysis, enforces secure-coding policy in CI/CD, identifies introduced vulnerabilities, and validates SDLC (Secure Software Development Life Cycle) compliance with PCI-DSS and SOX-style controls, closing the loop from development pipeline to operational posture \cite{chippagiri2025pci}.

\subsection{Bounded autonomy}\label{sec:hitl}

\begin{figure}[!t]
\centering
\includegraphics[width=1\columnwidth]{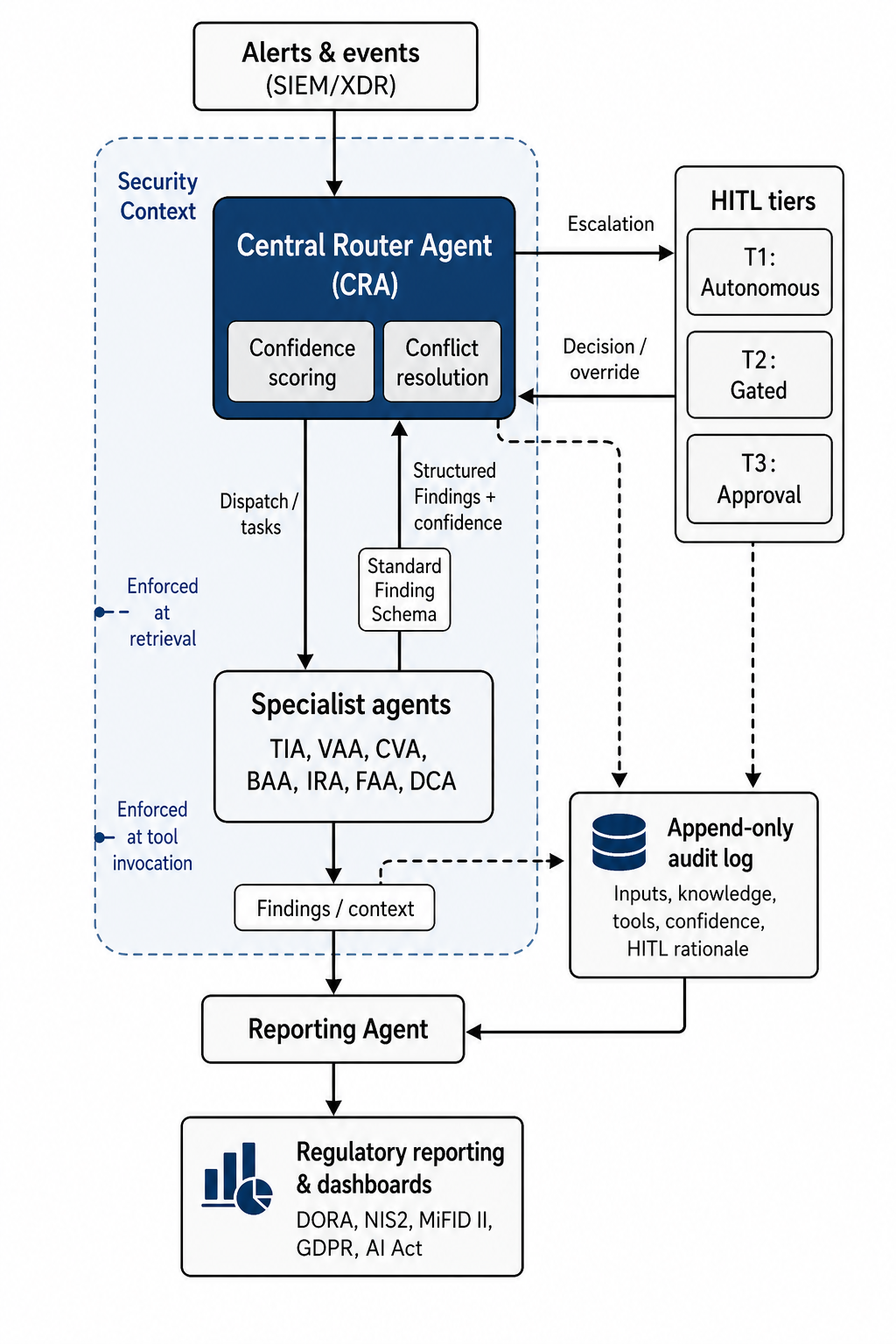}
\caption{Agent orchestration and trust envelope. The Security Context flows around the CRA and specialists; the CRA performs confidence scoring and conflict resolution; HITL tiers route decisions to autonomous, gated, or approval paths; every input, retrieval, tool call and confidence is written to an append-only audit log; the Reporting capability feeds regulatory dashboards.}
\label{fig:orchestration}
\end{figure}

As shown in Fig.~\ref{fig:orchestration}, three HITL tiers are wired into the CRA: \emph{Tier 1} (autonomous, reversible) for known-indicator blocking and triage; \emph{Tier 2} (confidence-gated escalation) for ambiguous cases evaluated against per-agent thresholds; and \emph{Tier 3} (explicit approval) for high-severity, low-reversibility actions.
Every decision is recorded with full provenance, directly implementing the human-oversight requirement of EU AI~Act Article~14 and the audit trail expected under DORA.

\section{Methods: Hybrid Foundations and Complementary Technologies}\label{sec:methods}

The platform is \emph{model-agnostic}: any sufficiently capable LLM backbone can serve as the inference layer, with private model deployment, for inference on infrastructure that preserves data residency under DORA, NIS2 and GDPR.
The remainder of this section details the methods stack.

\subsection{Hybrid integration with classical security tooling}\label{sec:methods-tools}

The platform consumes normalised security events from a generic SIEM/XDR over a messaging/alert system (streaming bus) and writes back into it~--~detection rules synthesised from agent-discovered patterns, alert enrichment with TTP and asset context, and adaptable, by the CRA agent, SOAR playbooks (agent skills)~--~closing the analytical loop without removing the deterministic detection floor that the SIEM provides.
Tool integration on the agent side is primarily through direct library bindings and CLI wrappers; throughput-sensitive paths (telemetry ingestion, containment commands, vulnerability scans, CI/CD hooks) require predictable latency, deterministic argument schemas and full audit, which are easier to guarantee through versioned CLI wrappers than through dynamically discovered MCP descriptors.
MCP is an optional external integration protocol behind the Tool Adapter Layer for cross-service exploratory investigations and for integrating external services that already expose MCP endpoints.

\subsection{Knowledge stack and partitioned RAG}\label{sec:methods-rag}

Each agent queries its own vector partition, scoped at retrieval time by the Security Context, against four classes of source: operational telemetry (system and application logs, file-integrity events, kernel telemetry, transaction streams, email and messaging, source-code and CI/CD events); external cyber threat intelligence (STIX/TAXII feeds, vendor advisories, exploit prediction scoring system (EPSS) and KEV\footnote{https://www.cisa.gov/known-exploited-vulnerabilities-catalog} scoring); authoritative knowledge bases (MITRE ATT\&CK, CVE/NVD, CWE, CAPEC, NIST CSF, NIST 800-53, ISO/IEC 27001); and regulatory corpora (DORA, NIS2, GDPR, PSD2, 4AML, AI Act, EBA and ECB guidelines).
Retrieval is hybrid (dense + sparse), augmented by graph traversal over CVE\,$\to$\,CWE\,$\to$\,CAPEC\,$\to$\,ATT\&CK chains.
Continuous CTI ingestion keeps the partitions current; vector stores are themselves treated as security-sensitive: encrypted at rest and in transit, integrity-monitored, and protected against poisoning attacks~\cite{ferrag2025prompt}.

\subsection{Federated cross-institutional knowledge sharing}\label{sec:methods-federated}

What is shared across participating institutions is \emph{accumulated agent knowledge} such as detection rules or workflows, identified behavioural patterns (e.g., impersonation attempts) and agent skills. These are aggregated under secure multi-party computation, secure aggregation and differential-privacy budgets calibrated to the sensitivity of the underlying data.
Smaller institutions inherit detection capability from the collective without exporting customer data. Thus,  each deployment becomes a contribution to a continuously refined collective agentic defence.

\subsection{Autonomous adversarial validation in digital twins}\label{sec:methods-pentest}

The platform can attach digital-twin replicas of the institution's critical infrastructure as an optional capability pack for safe, continuous adversarial validation.
LLM-driven red-team agents~\cite{ref-redteamllm} may operate inside the twin, attempting industry-specific attack patterns against payment systems, banking portals, trading APIs and credential flows.
Successful exploit paths can be converted into detection rules ingested by the SIEM, prioritisation hints for the VAA, and updated guardrails for the agents themselves, closing an otherwise asymmetric offensive--defensive loop in a controlled environment.

\subsection{eBPF-based kernel telemetry on financial endpoints}\label{sec:methods-ebpf}

Critical financial infrastructure is increasingly composed of edge devices: ATMs, POS terminals, mobile banking and payment apps, payment gateways, broker feeds.
The platform can consume eBPF (extended Berkeley Packet Filter) instrumentation through a telemetry provider for low-overhead syscall and network visibility on these endpoints, feeding the Behavioural and Forensic agents at sub-second latency without disturbing production workloads \cite{zhan2022shrinking}.
The combination of eBPF telemetry and LLM-assisted reasoning can enable behaviour-level detection of patterns that signature-based controls miss: lateral movement on POS networks, unusual administrative escalation on core banking servers, anomalous market-data feed handling.

\subsection{Quantum Token authentication}\label{sec:methods-quantum}

For high-value transaction flows the platform can integrate a \emph{quantum-token-based two-factor authentication (2FA)} provider, exploiting the no-cloning theorem \cite{schiansky2023demonstration}. It may be used in bank-to-bank and custodian--asset-manager transactions where each token is single-use, unforgeable, and any tampering is detectable by construction.

\subsection{Skill-based agent adaptation}\label{sec:methods-skills}

Each agent loads a base capability surface plus a set of \emph{skills}~--~composable, versioned, signed packages that bundle a prompt template, a tool binding, an RAG query pattern, a regulatory mapping and KPI targets.
Skills replace the monolithic system prompt with a context-activated, evolvable specialisation layer.
Examples: an \texttt{aml-eu-emoney} skill for the AML use case (Section~\ref{sec:uc-aml}); a \texttt{mifid-ii-incident} skill for HFT regulatory reporting (Section~\ref{sec:uc-hft}); a \texttt{pci-dss-v4} skill for the DCA.
We discuss skill-based adaptation as an open research direction in Section~\ref{sec:discussion}.

\section{Use Cases}\label{sec:usecases}

CyberAId will be validated at four representative use cases drawn from financial sub-sectors with materially different threat profiles, data sources, regulatory regimes, and tool integrations.

\subsection{UC-1: Tiered client impersonation detection (private banking)}\label{sec:uc-impersonation}

\begin{figure}[ht]
\centering
\includegraphics[width=1\columnwidth]{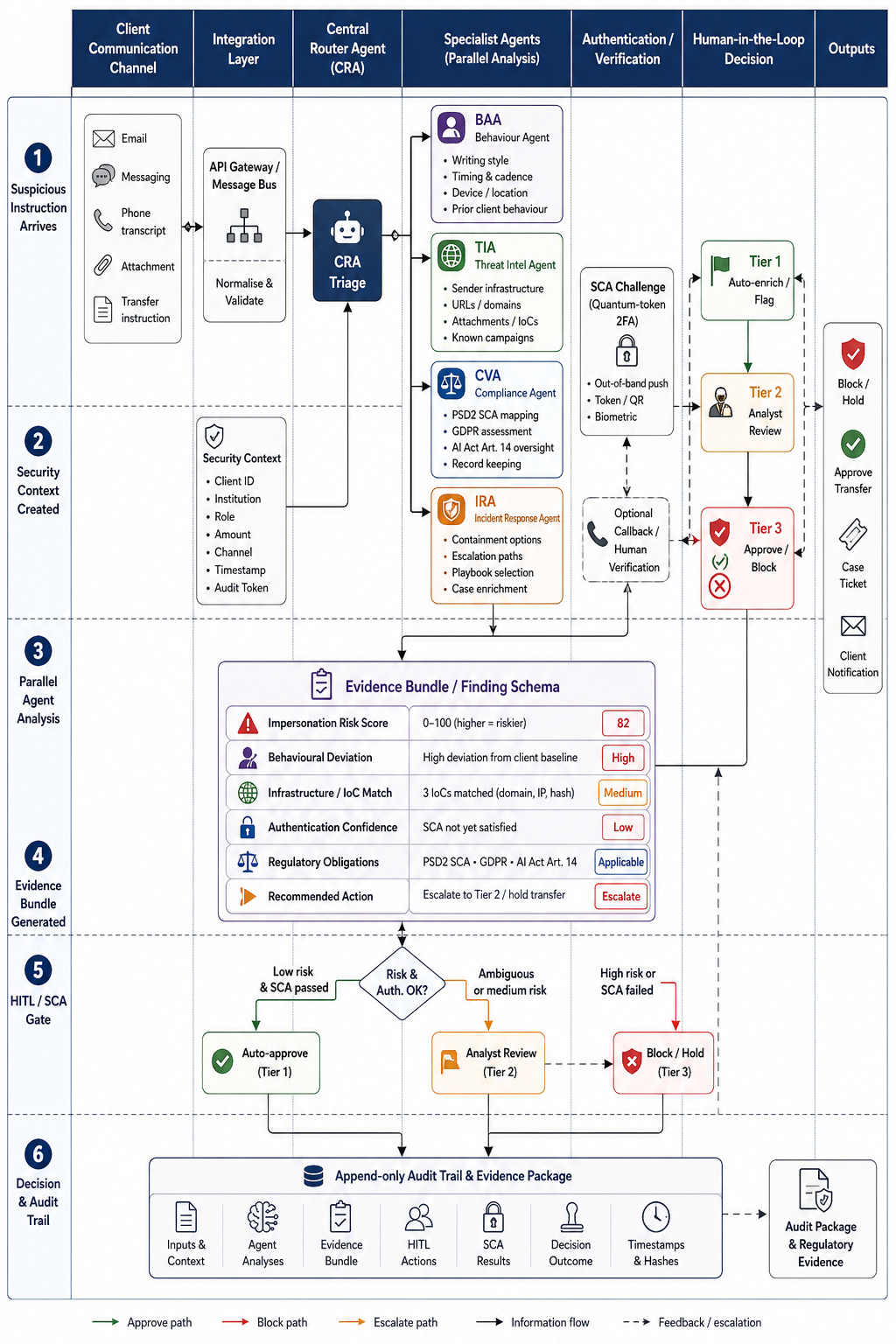}
\caption{Client-impersonation use case (UC-1). A suspicious order is wrapped in a Security Context, triaged by the CRA, and analysed in parallel by the Behavioural, Compliance, Vulnerability and Incident-Response agents; an optional quantum-token authentication challenge can gate high-value flows; an Evidence Bundle is assembled in the standard Finding Schema; HITL escalation is determined by confidence, value, and regulatory obligation; every decision is captured in an append-only audit log.}
\label{fig:uc-impersonation}
\end{figure}

Smaller asset managers routinely relay client orders to larger custodian institutions, creating opportunity for attackers to insert fraudulent instructions via compromised email accounts or endpoint devices~--~typically at amounts kept below verification thresholds.
The platform applies LLM-driven content analysis of order and transfer emails to detect linguistic deviation from the client's communication baseline, structural anomalies in headers and attachments, and inconsistencies between the message and the client's historical patterns.
Federated learning across custodians and asset managers shares detection improvements without exposing client data, so smaller institutions inherit large-bank detection capability.
An optional quantum-token-based 2FA channel can bind high-value instructions to a physically unforgeable token (Section~\ref{sec:methods-quantum}).
The end-to-end workflow is illustrated in Fig.~\ref{fig:uc-impersonation}.
Agents engaged: TIA, BAA, IRA, CVA.
Regulatory hooks: PSD2 strong customer authentication, GDPR, AI Act oversight on automated decision support.

\subsection{UC-2: Anti-money-laundering for payment service providers}\label{sec:uc-aml}

E-money institutions face a structural false-positive problem with rule-based AML, with detection rates above 95\% in name only and limited capacity to catch sophisticated structuring or layering~\cite{wu2024threatmodeling}.
The platform builds dynamic transaction graphs across e-wallet, IBAN-connected and card-payment ecosystems, applies graph analytics for detecting structuring, smurfing and layering, and uses the BAA and TIA jointly to identify shell-account and nominee-account patterns.
The Forensic Analysis Agent and the IRA can replay suspicious flows inside a digital twin capability pack (Section~\ref{sec:methods-pentest}) for safe analyst investigation.
The CVA generates suspicious-activity reports and audit packages aligned with 4AML and national reporting templates.
Federated knowledge sharing accelerates detection of cross-institutional laundering rings.
Agents engaged: BAA, TIA, FAA, IRA, CVA, CRA, Reporting.

\subsection{UC-3: Multi-channel anomaly detection and incident response (retail banking)}\label{sec:uc-retail}

Large retail banks operate at a scale~--~tens of thousands of servers, millions of clients, multi-petabyte log warehouses~--~at which alert volumes overwhelm even mature SOCs.
The platform can consume eBPF-enhanced telemetry across endpoints (Section~\ref{sec:methods-ebpf}), feed the BAA with kernel-level syscall and network signals, and use the CRA to perform pipelined triage at low latency.
Confirmed incidents trigger the IRA to execute pre-approved playbooks under tier-2 HITL, while the FAA preserves evidence in parallel.
The CVA maintains a live control dashboard and the Reporting capability drafts incident notifications within the 4-hour DORA window.
Agents engaged: BAA, TIA, VAA, IRA, FAA, CVA, CRA, Reporting.
Complementary technologies: eBPF, federated detection-rule sharing.

\subsection{UC-4: Cyber-resilient high-frequency trading}\label{sec:uc-hft}

High-frequency trading desks combine three high-value asset classes: proprietary algorithms (intellectual property), market-data feeds (integrity), and continuous availability under regulatory observation by national supervisors.
The platform monitors algorithm repositories and runtime behaviour with the DCA and the BAA for unauthorised access or model-output drift, validates market-data feeds for integrity and triggers established failover paths automatically, and uses the CRA-Reporting pair to synthesise forensic evidence and incident reports under DORA and MiFID~II.
Latency budget for any agent-side action is below second on the trading-critical path; the bulk of agent reasoning happens out-of-band on the monitoring path.
Agents engaged: DCA, BAA, VAA, IRA, FAA, CVA, CRA, Reporting.

\section{Discussion and Open Research Directions}\label{sec:discussion}

\subsection{Skill-based adaptation as a research direction}\label{sec:disc-skills}

Section~\ref{sec:methods-skills} introduced \emph{skills} as composable specialisation packages.
Conceptually, skills generalise three ideas already present in the agent literature: programmatic tool use~\cite{schick2023toolformer}, the ReAct-style separation of reasoning and acting~\cite{ref-react}, and plan-and-execute decomposition~\cite{wei2022chain,ref-plan-exec}, combined with modular orchestration patterns~\cite{ref-autogen}.
What is novel for the regulated cybersecurity setting is that skills must be \emph{provenance-bearing} (signed, versioned, attributable), \emph{conflict-resolvable} (when two skills disagree, the resolution must be auditable), and \emph{co-trainable} with the federated-learning loop of Section~\ref{sec:methods-federated}: aggregated agent knowledge and analyst feedback should refine the global skill library over time.

This raises concrete research questions: the semantics of \emph{skill conflict resolution} when, e.g., an AML skill flags a pattern that a PCI-DSS skill would suppress as a known compliance scan; \emph{skill poisoning}, since the threat model for externally authored skills extends the known RAG-poisoning surface; \emph{skill consistency under federated update}, including how differential-privacy budgets are allocated when each federated round refines multiple skills with overlapping evidence; and \emph{skill-level evaluation}, since existing benchmarks grade agents end-to-end rather than skill by skill.

\subsection{Other open directions}\label{sec:disc-open}

Three further gaps deserve attention before SOC-scale deployment becomes routine: \emph{production-grade evaluation}, since the gap between laboratory evaluation and operational deployment remains the single largest barrier to adoption; \emph{benchmark consolidation}, since available benchmarks offer overlapping coverage and incompatible scoring; and \emph{adversarial robustness of the agents themselves}, widely acknowledged as critical but rarely treated as a first-class concern.
CyberAId positions skill-based adaptation, federated knowledge sharing and digital-twin adversarial validation as the three operational levers most likely to convert these gaps into measurable progress. 

\section{Conclusion}\label{sec:conclusion}

Securing financial critical infrastructures with LLMs requires neither a smarter monolithic model nor a new SIEM but a \emph{hybrid} multi-agent system: specialist subagents reasoning over classical security tooling, sharing accumulated knowledge across institutions, and connecting to optional capability packs for adversarial validation, kernel-level visibility and quantum-token-based authentication.
CyberAId implements this position as a shared runtime with a Main Agent/CRA coordination layer, a Reporting capability, and specialist subagents under bounded human-in-the-loop autonomy, with skill-based adaptation as the most promising research direction toward continuously refined collective defence. 

\section*{Acknowledgment}

This work has received funding from the European Union under the DIGITAL EUROPE Programme, grant agreement No.~101249596 (CyberAId). \\
The authors used the AI tool Claude Opus 4.7 for grammar and style editing of the manuscript and Gemini NanoBanana to assist in figure generation, with all content reviewed and verified by the authors.

\bibliographystyle{IEEEtran}
\bibliography{references}

@article{ferrag2025prompt,
  title={From prompt injections to protocol exploits: Threats in LLM-powered AI agents workflows},
  author={Ferrag, Mohamed Amine and Tihanyi, Norbert and Hamouda, Djallel and Maglaras, Leandros and Lakas, Abderrahmane and Debbah, Merouane},
  journal={ICT Express},
  year={2025},
  publisher={Elsevier}
}

@article{wei2022chain,
  title={Chain-of-thought prompting elicits reasoning in large language models},
  author={Wei, Jason and Wang, Xuezhi and Schuurmans, Dale and Bosma, Maarten and Xia, Fei and Chi, Ed and Le, Quoc V and Zhou, Denny and others},
  journal={Advances in neural information processing systems},
  volume={35},
  pages={24824--24837},
  year={2022}
}

@article{schick2023toolformer,
  title={Toolformer: Language models can teach themselves to use tools},
  author={Schick, Timo and Dwivedi-Yu, Jane and Dess{\`\i}, Roberto and Raileanu, Roberta and Lomeli, Maria and Hambro, Eric and Zettlemoyer, Luke and Cancedda, Nicola and Scialom, Thomas},
  journal={Advances in neural information processing systems},
  volume={36},
  pages={68539--68551},
  year={2023}
}

@inproceedings{ref-autogen,
  title={Autogen: Enabling next-gen LLM applications via multi-agent conversations},
  author={Wu, Qingyun and Bansal, Gagan and Zhang, Jieyu and Wu, Yiran and Li, Beibin and Zhu, Erkang and Jiang, Li and Zhang, Xiaoyun and Zhang, Shaokun and Liu, Jiale and others},
  booktitle={First conference on language modeling},
  year={2024}
}

@misc{ref-attack,
  author       = {{MITRE Corporation}},
  title        = {{MITRE ATT\&CK Enterprise Matrix, Version 14}},
  howpublished = {\textit{MITRE ATT\&CK Enterprise Matrix, Version 14}},
  institution  = {MITRE},
  address      = {McLean, VA, USA},
  year         = {2023},
  note         = {Available online: \url{https://attack.mitre.org} (accessed on 1 April 2026)}
}

@article{drammeh2025multi,
  title={Multi-Agent LLM Orchestration Achieves Deterministic, High-Quality Decision Support for Incident Response},
  author={Drammeh, Philip},
  journal={arXiv preprint arXiv:2511.15755},
  year={2025}
}

@article{wu2024threatmodeling,
  title={ThreatModeling-LLM: Automating threat modeling using large language models for banking system},
  author={Wu, Tingmin and Yang, Shuiqiao and Liu, Shigang and Nguyen, David and Jang, Seung and Abuadbba, Alsharif},
  journal={arXiv preprint arXiv:2411.17058},
  year={2024}
}

@article{blefari2025cyberrag,
  title={CyberRAG: An agentic RAG cyber attack classification and reporting tool},
  author={Blefari, Francesco and Cosentino, Cristian and Pironti, Francesco Aurelio and Furfaro, Angelo and Marozzo, Fabrizio},
  journal={Future Generation Computer Systems},
  pages={108186},
  year={2025},
  publisher={Elsevier}
}

@inproceedings{ref-idsagent,
  title     = {IDS-Agent: An LLM Agent for Explainable Intrusion Detection in IoT Networks},
  author    = {Li, Yanjie and Xiang, Zhen and Bastian, Nathaniel D. and Song, Dawn and Li, Bo},
  booktitle = {NeurIPS 2024 Workshop on Open-World Agents: Synnergizing Reasoning and Decision-Making in Open-World Environments (OWA-2024)},
  year      = {2024},
  note      = {Poster},
}

@inproceedings{zhu2026teams,
  title={Teams of llm agents can exploit zero-day vulnerabilities},
  author={Zhu, Yuxuan and Kellermann, Antony and Gupta, Akul and Li, Philip and Fang, Richard and Bindu, Rohan and Kang, Daniel},
  booktitle={Proceedings of the 19th Conference of the European Chapter of the Association for Computational Linguistics (Volume 1: Long Papers)},
  pages={23--35},
  year={2026}
}

@article{ref-redteamllm,
  author    = {Brian Challita and Pierre Parrend},
  title     = {{RedTeamLLM}: An Agentic {AI} Framework for Offensive Security},
  journal   = {arXiv preprint arXiv:2505.06913},
  year      = {2025},
  eprint    = {2505.06913},
  archiveprefix = {arXiv},
  primaryclass  = {cs.CR},
}

@article{ref-llm-exploits,
  author    = {Richard Fang and Rohan Bindu and Akul Gupta and Daniel Kang},
  title     = {{LLM} Agents Can Autonomously Exploit One-Day Vulnerabilities},
  journal   = {arXiv preprint arXiv:2404.08144},
  year      = {2024},
  eprint    = {2404.08144},
  archiveprefix = {arXiv},
  primaryclass  = {cs.CR},
}

@inproceedings{ref-knowphish,
  author    = {Yuexin Li and Chengyu Huang and Shumin Deng and Mei Lin Lock and Tri Cao and Nay Oo and Hoon Wei Lim and Bryan Hooi},
  title     = {{KnowPhish}: Large Language Models Meet Multimodal Knowledge Graphs for Enhancing Reference-Based Phishing Detection},
  booktitle = {33rd USENIX Security Symposium (USENIX Security 24)},
  year      = {2024},
  publisher = {USENIX Association},
  address   = {Philadelphia, PA},
  pages     = {793--810},
  isbn      = {978-1-939133-44-1},
  url       = {https://www.usenix.org/conference/usenixsecurity24/presentation/li-yuexin},
}

@article{ref-soc-survey,
AUTHOR = {Srinivas, Siddhant and Kirk, Brandon and Zendejas, Julissa and Espino, Michael and Boskovich, Matthew and Bari, Abdul and Dajani, Khalil and Alzahrani, Nabeel},
TITLE = {AI-Augmented SOC: A Survey of LLMs and Agents for Security Automation},
JOURNAL = {Journal of Cybersecurity and Privacy},
VOLUME = {5},
YEAR = {2025},
NUMBER = {4},
ARTICLE-NUMBER = {95},
URL = {https://www.mdpi.com/2624-800X/5/4/95},
ISSN = {2624-800X},
DOI = {10.3390/jcp5040095}
}

@techreport{ref-ibm-breach,
  author      = {{IBM Security}},
  title       = {Cost of a Data Breach Report 2024},
  institution = {IBM},
  year        = {2024},
  address     = {Armonk, NY, USA},
  url         = {https://www.ibm.com/think/insights/cost-of-a-data-breach-2024-financial-industry},
  note        = {Accessed: 2026-04-01},
}

@inproceedings{ref-react,
  author    = {Shunyu Yao and Jeffrey Zhao and Dian Yu and Nan Du and Izhak Shafran and Karthik Narasimhan and Yuan Cao},
  title     = {{ReAct}: Synergizing Reasoning and Acting in Language Models},
  booktitle = {Proceedings of the International Conference on Learning Representations (ICLR 2023)},
  year      = {2023},
  address   = {Kigali, Rwanda},
}

@article{schiansky2023demonstration,
  title={Demonstration of quantum-digital payments},
  author={Schiansky, Peter and Kalb, Julia and Sztatecsny, Esther and Roehsner, Marie-Christine and Guggemos, Tobias and Trenti, Alessandro and Bozzio, Mathieu and Walther, Philip},
  journal={nature communications},
  volume={14},
  number={1},
  pages={3849},
  year={2023},
  publisher={Nature Publishing Group UK London}
}

@article{zhan2022shrinking,
  title={Shrinking the kernel attack surface through static and dynamic syscall limitation},
  author={Zhan, Dongyang and Yu, Zhaofeng and Yu, Xiangzhan and Zhang, Hongli and Ye, Lin},
  journal={IEEE Transactions on Services Computing},
  volume={16},
  number={2},
  pages={1431--1443},
  year={2022},
  publisher={IEEE}
}

@article{chippagiri2025pci,
  title={PCI DSS: A critical analysis of implementation, effectiveness, and legislative impact in payment card security},
  author={Chippagiri, Srinivas and Ramesh, Apoorva},
  year={2025}
}

@article{fatouros2026signal,
  title={Signal or Noise in Multi-Agent LLM-based Stock Recommendations?},
  author={Fatouros, George and Metaxas, Kostas},
  journal={arXiv preprint arXiv:2604.17327},
  year={2026}
}

@article{ref-plan-exec,
  title={A survey on large language model based autonomous agents},
  author={Wang, Lei and Ma, Chen and Feng, Xueyang and Zhang, Zeyu and Yang, Hao and Zhang, Jingsen and Chen, Zhiyuan and Tang, Jiakai and Chen, Xu and Lin, Yankai and others},
  journal={Frontiers of Computer Science},
  volume={18},
  number={6},
  pages={186345},
  year={2024},
  publisher={Springer},
  doi={10.1007/s11704-024-40231-1}

}

@article{ref-cortex,
  author    = {Bowen Wei and Yuan Shen Tay and Howard Liu and Jinhao Pan and Kun Luo and Ziwei Zhu and Chris Jordan},
  title     = {{CORTEX}: Collaborative {LLM} Agents for High-Stakes Alert Triage},
  journal   = {arXiv preprint arXiv:2510.00311},
  year      = {2025},
  eprint    = {2510.00311},
  archiveprefix = {arXiv},
  primaryclass  = {cs.CR},
}

@article{ref-survey-llm-sec,
  author    = {Feng He and Tianqing Zhu and Dayong Ye and Bo Liu and Wanlei Zhou and Philip S. Yu},
  title     = {The Emerged Security and Privacy of {LLM} Agent: A Survey with Case Studies},
  journal   = {ACM Computing Surveys},
  year      = {2025},
  doi       = {10.1145/3773080},
}

@techreport{ref-eba-ai,
  author    = {{European Banking Authority}},
  title     = {Report on Artificial Intelligence in the Banking Sector},
  institution = {EBA},
  year      = {2024},
  address   = {Paris, France},
  url       = {https://www.eba.europa.eu},
}

@misc{ref-dora,
  author    = {{European Parliament and Council}},
  title     = {Regulation ({EU}) 2022/2554 on Digital Operational Resilience for the Financial Sector ({DORA})},
  journal   = {Official Journal of the European Union},
  year      = {2022},
  address   = {Brussels, Belgium},
}

@misc{ref-nis2,
  author    = {{EU}},
  title     = {Directive ({EU}) 2022/2555 on Measures for a High Common Level of Cybersecurity Across the Union ({NIS2})},
  journal   = {Official Journal of the European Union},
  year      = {2022},
  address   = {Brussels, Belgium},
}

@misc{ref-aiact,
  author    = {{European Parliament and Council}},
  title     = {Regulation ({EU}) 2024/1689 Laying Down Harmonised Rules on Artificial Intelligence ({AI Act})},
  journal   = {Official Journal of the European Union},
  year      = {2024},
  address   = {Brussels, Belgium},
}

@inproceedings{ref-ctikg,
  author    = {Liangyi Huang and Xusheng Xiao},
  title     = {{CTIKG}: {LLM}-Powered Knowledge Graph Construction from Cyber Threat Intelligence},
  booktitle = {Proceedings of the First Conference on Language Modeling (COLM 2024)},
  year      = {2024},
  address   = {Philadelphia, PA, USA}
}

@article{ref-iris,
  author    = {Ziyang Li and Saikat Dutta and Mayur Naik},
  title     = {{IRIS}: {LLM}-Assisted Static Analysis for Detecting Security Vulnerabilities},
  journal   = {arXiv preprint arXiv:2405.17238},
  year      = {2024},
  eprint    = {2405.17238},
  archiveprefix = {arXiv},
  primaryclass  = {cs.CR},
  note      = {Accepted at ICLR 2025},
}

\end{document}